\documentclass{esannV2}

\usepackage{amssymb,amsmath,amsthm,array}

\usepackage{setspace}
\usepackage{nicefrac}
\usepackage[labelfont=bf,font=footnotesize]{caption}
\usepackage{color}
\usepackage{graphicx}
\usepackage{hyperref}
\usepackage{enumitem}
\usepackage{booktabs}
\theoremstyle{definition}

\begin{document}
\title{
In-Context Learning and Fine-Tuning GPT for Argument Mining
}

\author{J\'{e}r\'{e}mie Cabessa$^1$, Hugo Hernault$^2$ and Umer Mushtaq$^3$
%
%
\vspace{.3cm}\\
%
$^1$DAVID Lab, UVSQ -- University Paris-Saclay, 78035 Versailles. France \\
$^2$Playtika Ltd., CH-1003 Lausanne, Switzerland \\
$^3$L3i, University of La Rochelle, 17042 La Rochelle, France
}

\maketitle

\begin{abstract}
Large Language Models (LLMs) have become ubiquitous in NLP and deep learning. In-Context Learning (ICL) has been suggested as a bridging paradigm between the training-free and fine-tuning LLMs settings. In ICL, an LLM is conditioned to solve tasks by means of a few solved demonstration examples included as prompt. Argument Mining (AM) aims to extract the complex argumentative structure of a text, and Argument Type Classification (ATC) is an essential sub-task of AM. We introduce an ICL strategy for ATC combining $k$NN-based examples selection and majority vote ensembling. 
In the training-free ICL setting, we show that GPT-4 is able to leverage relevant information from only a few demonstration examples and achieve very competitive classification accuracy on ATC. We further set up a fine-tuning strategy incorporating well-crafted structural features given directly in textual form. In this setting, GPT-3.5 achieves state-of-the-art performance on ATC. Overall, these results emphasize the emergent ability of LLMs to grasp global discursive flow in raw text in both off-the-shelf and fine-tuned setups.
\end{abstract}

\section{Introduction}
\label{Introduction}

Large Language Models (LLMs) have become ubiquitous in Deep Learning and have shown impressive capabilities in most NLP tasks~\cite{ZhaoEtAl23}. 
In the main, LLMs are used in two distinct settings: (i) training-free, where the pre-trained LLM is used for inference without any parameter adjustment, and (ii) fine-tuning, where the parameters of the LLM are updated through supervised training to enable transfer learning on a downstream task.

Recently, In-Context Learning (ICL) has been suggested as a bridging pa\-ra\-digm between the training-free and fine-tuning settings. ICL is a prompt engineering technique whereby an LLM is conditioned to solve tasks by means of a few solved demonstration examples included as part of its input prompt~\cite{DongEtAl23}. Generally, the input prompt includes task instructions, the current input sample to be solved as well as several solved input-output pair examples. In this way, ICL maintains the training-free posture (parameters frozen) of the LLM while at the same time providing it with some supervision through demonstration examples. It also enables direct incorporation of selected features inside the prompt template, thereby obviating the need for architecture customization. Creative ICL strategies combining $k$NN-based examples selection, generated chain-of-thought (CoT) prompting, and majority vote ensembling have been proposed and shown to outperform fine-tuning approaches~\cite{Microsoft23,CoTWeiEtAl2023,LeiEtAl23,WangSelfConsistency23}. In the main, $k$NN-based examples selection optimizes the process of learning from few examples and ensembling increases the robustness of the predictions~\cite{CapabilitiesNori23,Microsoft23,LeiEtAl23}.

Argument Mining (AM) focuses on extracting the underlying argumentative and discursive structure from raw text~\cite{AMPalauMoens09}. Argument Type Classification (ATC), which involves classifying argumentative units in text according to their argumentative roles, is the crucial sub-task in AM. Research has shown that the argumentative role of a unit cannot be inferred solely for its text: additional structural and contextual information is needed~\cite{ParsingStabGurevych17}. This additional information can be incorporated via feature engineering~\cite{ParsingStabGurevych17}, memory-enabled neural architectures~\cite{PotashEtAl17,SpanRepKuribayashiEtAl19} or LLM-based hybrid methods~\cite{BERTFeaTxtMushtaqCabessa22,BERTMinusMushtaqCabessa23}.

We introduce an ICL strategy for ATC combining $k$NN-based examples selection and majority vote ensembling. We experimented with various prompting templates in order to reveal the proper contribution of different contextual elements. In this training-free setting, we show that GPT-4~\cite{GPT42023} is able to leverage relevant information from only a few demonstration examples in order to achieve very competitive classification accuracy. We further set up a fine-tuning strategy incorporating well-crafted structural features input directly in textual form. In this setting, GPT-3.5 achieves state-of-the-art performance on the ATC task. Overall, these results emphasize the emergent ability of LLMs to grasp global discursive flow in raw text in both off-the-shelf and fine-tuned setups~\cite{wei2022emergent}. The code is available on GitHub \href{https://github.com/JeremCab/AMwithGPT}{here}.

\section{Methodology}
\label{Methodology}

\subsection{Dataset}
\label{Dataset}

We consider the Persuasive Essays (PE) dataset which consists of 402 structured essays on various topics~\cite{ParsingStabGurevych17}. The train and test sets are composed of 322 and 80 essays, respectively. The statistics of the PE dataset are given in Table~\ref{table:datasetstats}. The argument type classification (ATC) task consists of predicting the type of each argument component (AC): `Major Claim', `Claim' or `Premise'.

\renewcommand{\arraystretch}{0.9}
\begin{table}[h!]
    \centering
    {\footnotesize   
    \begin{tabular}{*{4}{l}}
    \toprule
\multicolumn{2}{c}{{\bf Corpus Statistics}} & \multicolumn{2}{c}{{\bf Component Statistics}} \\
    \midrule
    Tokens  & 147,271 &  Major Claims  & 751 \\
    Sentence & 7,116 & Claims & 1,506 \\
    Paragraphs & 1,833 & Premises & 3,832 \\
    Essays & 402 & Total & 6,089 \\    
    \bottomrule
    \end{tabular}
    \caption{PE dataset statistics.}
    \label{table:datasetstats}
    }
\end{table}

\subsection{In-Context Learning (ICL)}
\label{In-Context Learning (ICL)}

{\it In-context learning (ICL)} refers to the emergent ability of LLMs to solve a task based on a few demonstration examples given as contextual information~\cite{DongEtAl23}. As the major advantage, the ICL paradigm does not require any further fine-tuning of the model's parameters (i.e.~training-free framework). Formally, let $x$ be a query input text, $Y = \{ y_1, \dots, y_k \}$ be a set of candidate answers, and $C = \left[I; t(x_1, y_{i_1}); \dots; t(x_k, y_{i_k})\right]$ be a context composed of instructions $I$ concatenated with text representations of example input-output pairs $(x_j , y_{i_j})$. Then, the LLM $\mathcal{M}$ predicts the answer $\hat y$ such that 
$
\hat y = \arg \max_{y_i \in Y} P_{\mathcal{M}} (y_i \mid C; x) \,,
$
where $P_{\mathcal{M}} (y_i \mid C; x)$ is the probability that $\mathcal{M}$ generates $y_i$ when $C$ and $x$ are given as prompt. The main rationale behind ICL is that the consideration of a well-chosen context $C$ increases the probability of $\mathcal{M}$ predicting the correct answer $y$ for input $x$, i.e. $P_{\mathcal{M}} (y \mid C; x) > P_{\mathcal{M}} (y \mid x)$.

\begin{figure}[h!]
\begin{center}
\includegraphics[width=0.8\textwidth]{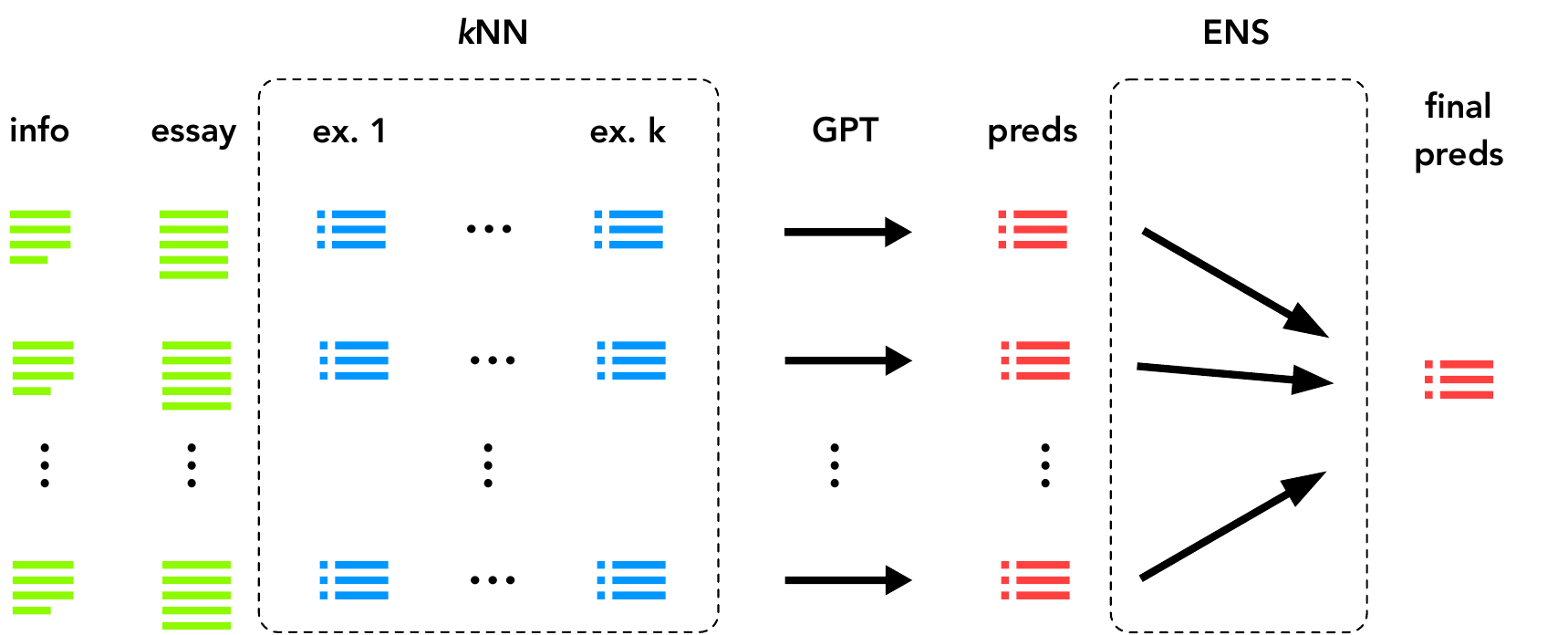}
\caption{2-step ICL approach: a $k$NN-based example prediction ($k=3, 5$) step followed by an $n$-Ensembling ($n=3, 5$) step (cf.~text for further details). 
}
\label{ICL}
\end{center}
\end{figure}

We consider a 2-step ICL strategy for argument type classification (ATC) inspired by a recent study~\cite{Microsoft23} (see Figure~\ref{ICL}). More precisely, let $E$ be an essay containing argument components (ACs) $c_1,\dots,c_m$ with their corresponding true classes $y_1, \dots, y_m$, where each $y_i \in \{ \text{Claim}, \text{Major Claim}, \text{Premise} \}$. 
The LLM generates class predictions $\hat y_1, \dots, \hat y_m$ for $c_1,\dots,c_m$ as follows:
\begin{enumerate}[itemindent=0pt,leftmargin=20pt,topsep=3pt,parsep=0pt,label=(\arabic*)]
\item {\bf $k$NN-based examples selection ($k=3, 5$):} First, $N$ neighboring essays $E_1, \dots, E_N$ of $E$ are selected according to some similarity measure. The selection of $E_1, \dots, E_N$ involves three options: (i) random selection ({\bf $k$RN}), (ii) selection based on essays having the same (or the closest) number of ACs as $E$ ({\bf $k$NN$^\text{len}$}) and (iii) selection based on essays whose title embeddings (with OpenAI's \texttt{text-embedding-ada-002}) are closest with respect to cosine similarity to the title embedding of $E$ ({\bf $k$NN}). Then, $k = \nicefrac{N}{2}$ essays, $E_{i_1}, \dots ,E_{i_k}$, are randomly chosen from $E_1, \dots, E_N$. Afterwards, a prompt containing all the ACs and their corresponding classes in these $k$ essays is constructed. Finally, the LLM predicts the classes $\hat y_1, \dots, \hat y_m$ of $c_1, \dots, c_m$ based on this prompt. 
\item {\bf $n$-Ensembling ($n=3, 5$):} The $k$NN-based examples selection step, which involves randomness, is repeated $n$ times ({\bf $n$Ens}), leading to a set of $n$ class predictions sequences $\{ (\hat y_{i,1}, \dots, \hat y_{i,m}) : i = 1,\dots n \}$. The final class predictions $\hat y_1, \dots, \hat y_m$ of $c_1, \dots, c_m$ are obtained by applying a component wise majority vote to the $n$ predictions sequences. 
\end{enumerate}

According to this ICL strategy, the classes $\hat y_1, \dots, \hat y_m$ are predicted all-at-once (see Figure~\ref{ICL}). A similar ICL strategy where the classes $\hat y_1, \dots, \hat y_m$ are inferred one-by-one (i.e.~each model inference leads to a single prediction $\hat y_j$) has been considered, but shown to be significantly less efficient. Due to space constraints, the latter results are omitted in this work.

\subsection{Additional information and features}

To help the LLM in its predictions, complementary information about the task is generally added in the prompt. Here, the definitions of the classes (cf.~\href{https://tudatalib.ulb.tu-darmstadt.de/handle/tudatalib/2422}{guidelines}) together with their statistics in the train set can be included in the prompt ({\bf info}). Similarly, the whole text of the essay can be added in the prompt ({\bf essay}).

The text of an argument component alone does not suffice to predict its argumentative class. In fact, the addition of contextual, structural and syntactic features is indispensable for reaching at least satisfactory prediction accuracy~\cite{ParsingStabGurevych17}. Based on this insight, an approach where contextual, structural and syntactic features are given {\it as text} (instead of numerically) to the LLM has been proposed~\cite{BERTFeaTxtMushtaqCabessa22,BERTMinusMushtaqCabessa23}. Here, structural features of the ACs can be added in the prompt in the following textual form ({\bf fts}): {\it Is the AC first in its paragraph: yes/no. Is the AC last in its paragraph: yes/no. Is the AC in the introduction of the essay: yes/no. Is the AC in the conclusion of the essay: yes/no.}

\subsection{Fine-tuning}
\label{LLM_finetuning}

{\it Fine-tuning} refers to the process of further training a pre-trained LLM on a downstream task. 
We fine-tune GPT--3.5--turbo with and without injected textual features (GPT--3.5--FeaTxt and GPT--3.5). As contextual features, we include the title of the essay and the complete sentence in which the argument component appears. As structural features, we include the paragraph number as well as indications of whether the argument component is present in the introduction/conclusion paragraph and whether it is the first and last component in its paragraph, in the form given above. 

%
%
%
%
%

\section{Results}
\label{Results}

We used GPT--4 and GPT--3.5 on the ATC task with the 2-step ICL strategy described in Section~\ref{In-Context Learning (ICL)}. We experimented with various prompting configurations in order to assess the proper contribution of the different contextual elements. The results are reported in Table~\ref{table_ICL}.

Our 2-step ICL approach utilizing $5$NN and $5$-ensembling achieves a macro F1-score of 83.6\% (Line $5$). These results surpass those attained by the fine-tuned BERT--FeaTxt and BERT--MINUS--FeaTxt models (see Table~\ref{table_FT})~\cite{BERTFeaTxtMushtaqCabessa22,BERTMinusMushtaqCabessa23}. Therefore, our ICL approach is capable of capturing the global argumentative flow of the essays through relevant contextual information and demonstration examples, without requiring fine-tuning. Note that these results depend on the complexity of the LLM, as GPT--3.5 with the very same prompting configuration obtains markedly worse performance than GPT--4 (line $7$). This phenomenon possibly points to an emergent ability acquired through model scaling.

In addition, lines $2$ and $4$ show that our two examples selection processes ($5$NN$^\text{len}$ vs $5$NN) yield comparable results. 
Lines $1$ and $2$ emphasize the added value of the ensembling process, which increases the robustness of the predictions.
Lines $3$ and $5$ show the strong ability of GPT to leverage additional information (class definitions and statistics). 
But not all information is necessarily beneficial. In fact, although structural features are known to be crucial for the ATC task~\cite{ParsingStabGurevych17}, lines $5$ and $6$ reveal that adding them explicitly turns out to be counter-productive. Our rationale is the the model is better able to infer these features from the raw argumentative flow of the demonstration examples.

\begin{table}[h!]
    \centering
    {\footnotesize
    \begin{tabular}{l*{5}{c}}
    \toprule
    {\bf Prompt} & {\bf MC} & {\bf C} & {\bf P} & {\bf F1}  \\
    \midrule
    (GPT--4) &&& \\
    info + essay \textcolor{white}{+ fts} + 5NN$^\text{len}$ 	& 0.894 & 0.681 & 0.896 & 0.823 \\
    info + essay \textcolor{white}{+ fts} + 5NN$^\text{len}$ + 3Ens	& 0.900 & 0.692 & 0.898 & 0.830 \\
    \textcolor{white}{info +} essay \textcolor{white}{+ fts} + 5NN\textcolor{white}{$^\text{len}$} + 5Ens				& 0.905 & 0.686 & 0.890 & 0.827 \\
    info + essay \textcolor{white}{+ fts} + 5NN\textcolor{white}{$^\text{len}$} + 3Ens			& 0.910 & 0.688 & 0.893 & 0.831 \\
    info + essay \textcolor{white}{+ fts} + 5NN\textcolor{white}{$^\text{len}$} + 5Ens			& 0.914 & 0.698 & 0.897 & {\bf 0.836} \\
    info + essay + fts + 5NN\textcolor{white}{$^\text{len}$} + 5Ens		& 0.895 & 0.671 & 0.897 & 0.821 \\
    \midrule
    (GPT--3.5) &&& \\
    info + essay + \textcolor{white}{+ fts} + 5NN\textcolor{white}{$^\text{len}$} + 5Ens			& 0.840 & 0.594 & 0.877 & 0.770 \\
    \bottomrule
    \end{tabular}
    \caption{Results for the ATC task using the 2-step ICL strategy.}
    \label{table_ICL}
    }
\end{table}


We also fine-tuned GPT--3.5--turbo for $1$ and $2$ epochs, with and without including the structural features described in previous works~\cite{BERTFeaTxtMushtaqCabessa22,BERTMinusMushtaqCabessa23}. The results are reported in Table~\ref{table_FT}. 
In line with previous works~\cite{ParsingStabGurevych17}, the addition of structural features significantly improves GPT's performance (lines $7$-$9$). Most importantly, the fine-tuned GPT--3.5--FeaTxt achieves a macro F1-score of $0.863$ which outperforms the SoTA ($0.857$) for the single ATC task~\cite{SpanRepKuribayashiEtAl19}. This shows that the fine-tuning approach outperforms the ICL approach (see Tables~\ref{table_ICL} and \ref{table_FT}). This suggests that, as opposed to simpler tasks~\cite{Microsoft23}, ATC might not be solvable optimally without resorting to further task-specific learning.


\begin{table}[h!]
    \centering
    {\footnotesize
    \begin{tabular}{l*{5}{c}}
    \toprule
    {\bf Model} & {\bf MC} & {\bf C} & {\bf P} & {\bf F1}  \\
    \midrule
    Stab and Gurevych~\cite{ParsingStabGurevych17} & 0.891 & 0.682 & 0.903 & 0.826 \\
    Potash et al.~\cite{PotashEtAl17} & 0.894 & 0.732 & 0.921 & 0.849 \\
    Kuribayashi et al.~\cite{SpanRepKuribayashiEtAl19} & & & & 0.856 \\
    \midrule
    BERT (4 ep.)~\cite{BERTFeaTxtMushtaqCabessa22} & 0.703  & 0.507 &  0.841  & 0.686 \\
    BERT--FeaTxt (4 ep.)~\cite{BERTFeaTxtMushtaqCabessa22} & 0.855 & 0.711 & 0.908 & 0.826 \\    
       BERT--MINUS--FeaTxt (4 ep.)~\cite{BERTMinusMushtaqCabessa23} & 0.900  & 0.687 & 0.903  & 0.831 \\
    \midrule
    GPT--3.5 (1 ep.) & 0.524 & 0.416 & 0.795 & 0.578 \\
    GPT--3.5--FeaTxt (1 ep.) & 0.905 & 0.745 & 0.915 & 0.855 \\
    GPT--3.5--FeaTxt (2 ep.) & 0.934 & 0.740 & 0.914 & \textbf{0.863} \\
    \bottomrule
    \end{tabular}
    \caption{Results of the ATC task achieved by previous non-LLM methods~\cite{ParsingStabGurevych17,PotashEtAl17,SpanRepKuribayashiEtAl19}, by fine-tuned BERT-based models~\cite{BERTMinusMushtaqCabessa23}) and by fine-tuned GPT--3.5 and GPT--3.5--FeaTxt (ours). 
    }
    \label{table_FT}
    }
\end{table}

\section{Conclusion}
\label{Conclusion}

We introduce a 2-step in-context learning (ICL) strategy for argument type classification (ATC). Our strategy applied to GPT--4 implements training-free learning by combining $k$NN-based examples selection and majority vote-based ensembling mechanisms. To leverage the inherent transfer learning capabilities of LLMs, we also fine-tune GPT--3.5 with and without selected well-crafted features. Overall, our results confirm that the argument type of a component can only be optimally captured through the discursive flow of the corpus text, either given explicitly (ICL approach) or encoded into selected features (fine-tuning approach). For future work, we plan to apply the ICL strategy to all the Argument Mining sub-tasks so as to achieve an LLM-based, training-free, end-to-end AM pipeline.

\bibliographystyle{unsrt}
\bibliography{bibliography}



\end{document}